\documentclass{imsart}

\RequirePackage{amsthm,amsmath,amsfonts,amssymb}
\RequirePackage[numbers]{natbib}
\RequirePackage[colorlinks,citecolor=blue,urlcolor=blue]{hyperref}
\RequirePackage{graphicx}

\usepackage[utf8]{inputenc} 
\usepackage[T1]{fontenc}    
\usepackage{hyperref}       
\usepackage{url}            
\usepackage{booktabs}       
\usepackage{amsfonts}       
\usepackage{nicefrac}       
\usepackage{microtype}      
\usepackage{amssymb}        
\usepackage[disable]{todonotes} 

\usepackage{changepage} 
\usepackage{algorithm} 
\usepackage[noend]{algpseudocode} 

\usepackage{enumerate} 

\begin{document}

\begin{frontmatter}

\title{Tree pyramidal adaptive importance sampling}
\runtitle{Tree pyramidal adaptive importance sampling}

    
\begin{aug}
    \author[A]{\fnms{Javier} \snm{Felip}}
    \and
    \author[A]{\fnms{Nilesh} \snm{Ahuja}}
    \and
    \author[A]{\fnms{Omesh} \snm{Tickoo}}
    \address[A]{Intel Labs, Intel Corporation}
\end{aug}

\begin{abstract}
This paper introduces Tree-Pyramidal Adaptive Importance Sampling (TP-AIS), a novel
iterated sampling method that outperforms state-of-the-art approaches 
like deterministic mixture population Monte Carlo (DM-PMC \cite{DM-PMC}), mixture population 
Monte Carlo (M-PMC \cite{M-PMC}) and layered adaptive importance sampling (LAIS \cite{LAIS}).

TP-AIS iteratively builds a proposal distribution parameterized by a tree pyramid, where each
tree leaf spans a subspace that represents its importance density. After
each new sample operation, a set of tree leaves are subdivided for improving the approximation of 
the proposal distribution to the target density. Unlike the rest of the methods in the literature,
TP-AIS is parameter free and requires no tuning to achieve its best performance.

We evaluate TP-AIS with different complexity randomized target probability density 
functions (PDF) and also analyze its application to different dimensions. The results are 
compared to state-of-the-art iterative importance sampling approaches and other baseline MCMC 
approaches using Normalized Effective Sample Size (N-ESS), Jensen-Shannon Divergence, and 
time complexity.
\end{abstract}

\end{frontmatter}


\section{Introduction}
\label{sec:intro}
A central task in the application of probabilistic models is the estimation of latent or unknown variables from observed noisy data. Within the Bayesian framework, this involves combining prior knowledge about the latent variables with the information in the observations to obtain a joint posterior probability distribution. Inference using such models typically involves evaluating queries such as finding values that maximize the posterior (MAP queries), or computing the posterior probability of some variables given evidence on the others (conditional queries). Often, it involves computing expectations of some function of interest with respect to the posterior. In most real-world applications, exact inference is not feasible either because the dimensionality of the latent space is too high or because the posterior distribution has a highly complex form for which expectations are not analytically tractable.

In such situations, we resort to approximate inference. Historically, sampling methods - also known as Monte Carlo (MC) methods - have been the method of choice for such problems. The goal of MC methods is to generate random numbers from a target distribution of interest, $\pi(x)$. These samples serve as an approximate representation of $\pi(x)$ and can be used to compute approximations to desired quantities such as expectations. Sampling methods are asymptotically exact in that they can generate exact results given infinite computational resources. Within the rich variety of sampling-based methods, a particularly important class are the \emph{Markov Chain Monte Carlo} (MCMC) methods. In these, samples are not drawn directly from $\pi(x)$ but instead from a Markov process whose stationary distribution is equal to $\pi(x)$. These methods have the elegant property that (under reasonable constraints) the state-distribution of the Markov process converges to $\pi(x)$ irrespective of the starting distribution. Further, they scale very well with the dimensionality of the sample space. 

However, there are often significant challenges in practical implementation. One is that the samples are generated sequentially and hence these methods cannot be easily parallelized unless some underlying structure in either the distribution or the transition mechanism is assumed. More serious are the problems of \emph{burn-in} and \emph{mixing}. Burn-in time refers to the number of steps we must wait before being able collect samples from the chain. This happens because the initial state distribution is arbitrary and we need to wait until it comes close to $\pi(x)$. Mixing time of a Markov chain is the time until the Markov chain is close to $\pi(x)$. Long mixing times occur in multi-modal distributions where the regions between modes are low-probability. In such situations, the chain might have explored one region well, but it can take a long time for it to transition between the modes, \cite{rubinstein_2008}.

An alternative to MCMC is the class of importance sampling (IS) methods. Historically, IS has been used to approximate expectations of the form $E[f(x)]=\int f(x)\pi(x)dx$, rather than generating samples from $\pi(x)$. Samples are drawn from an alternate distribution, $q(x)$, known as the \emph{proposal distribution} from which it is easier to generate samples. Each sample is assigned an \emph{importance weight} to correct the bias introduced by sampling from the wrong distribution. Unlike MCMC methods, there is no phenomenon of burn-in; the sample generation process can be parallelized; and all the generated samples along with their weights are retained, \cite{Owen2013}. The samples with their weights represent an approximation of the target distribution. The key challenge lies in choosing good proposal distributions. For poor choices of $q(x)$, the set of importance weights may be dominated by a few weights having large values, resulting in an effective sample size much smaller than the apparent sample size. This situation gets exacerbated at higher dimensions and can result in an exponential blowup in the number of samples required, resulting in very poor sampling efficiency.

We present here, therefore, the Tree Pyramid Adaptive Importance Sampling (TP-AIS) method that retains the desirable properties of Importance sampling (parallelizability, no burn-in) while achieving a higher sampling efficiency relative to state-of-the-art IS methods. We use a hierarchical data-structure called Tree Pyramids to describe the proposal distribution $\mathcal{Q}$ over the input probability space. Each node represents a K-dimensional subspace, with nodes further down the tree representing increasingly finer subspaces. Our algorithm adaptively divides the input space in a manner such that more samples are used to represent regions of higher probability density, while fewer samples are used for lower density regions. This enables us to efficiently approximate the target distribution $\pi$.

Unlike other state-of-the-art IS methods, TP-AIS is completely parameter-free and has anytime property that enables to make trade-offs between computation time and approximation quality.


The paper is structured as follows: Section~\ref{sec:related} provides an overview of sampling methods focusing on importance sampling algorithms related to our contribution and comments on their differences. Section~\ref{sec:methods} details the proposed sampling algorithm. Section~\ref{sec:evaluation} describes the methods used for evaluation and Section~\ref{sec:results} shows the results and discussion. The paper concludes with the final remarks and future directions in Section~\ref{sec:conclusion}.


\section{Related work}
\todo{NA: Review to define performance and sampling efficiency and ESS.}
\label{sec:related}
Sampling methods rely on the law of large numbers to guarantee asymptotic convergence to the exact solution. The rate of convergence depends on the algorithm and the problem at hand, and often it is one of the factors that determines the selection of the sampling method to use for a specific application.

\subsection{Direct sampling and MCMC}
When direct sampling from the target distribution is possible, a common option is to draw a set of samples and use them to build Monte Carlo approximations for the quantities of interest or a Kernel Density Estimate (KDE) that approximates the posterior. Sometimes, the target joint distribution $\pi(x_0...x_n)$ can be represented as a product of factors. If these factors represent conditional distributions $\pi(x) = \prod_{i=0}^n\pi(x_{i}|x_{pa_i})$, where $x_{pa_i}$ is the parent node of $x_i$, then it is possible to sample sequentially from the conditional distributions using an approach known as \emph{ancestral sampling} \cite{bishop06}. For more general factors, approaches such as \emph{Gibbs sampling} can be adopted \cite{Geman84}. As the complexity of the models increases, direct sampling might become challenging. For these cases, two closely related approaches exist: \emph{importance sampling} and \emph{approximate sampling}.

The performance of IS methods for high dimensional problems is not good and \emph{Monte Carlo Markov Chain} (MCMC) or \emph{variational inference} (VI) methods are used instead \cite{Geman84, Hastings70, hoffman14, duane87, jaakkola01}. Although in this paper we do not focus on those methods, one of our evaluation baselines is based on the \emph{MCMC Metropolis-Hastings} algorithm \cite{Hastings70}, and some adaptive IS methods include MCMC steps in their adaptation steps \cite{LAIS}. Among other problems, MCMC does not approximate the partition function and VI does not deal well with multi-modal target distributions.

\subsection{Multi-adaptive importance sampling}
As will be discussed in Section~\ref{sec:methods}, the selection of the proposal distribution has a big impact on the algorithm performance. The criteria for proposal distribution selection leads to different IS methods. A well-known method is \emph{rejection sampling} and its generalizations \cite{casella04}. Although it is an exact simulation method, its rejection mechanism requires many more sampling operations to produce a useful amount of samples (low-variance estimates) than other methods.

\emph{Multi-IS} methods use multiple proposal distributions to sample \cite{veach97}. Moreover, the different mixture components can be weighted and their weights tuned depending on previous samples \cite{owen00}. IS is naturally parallel, but in practice its implementations are sequential. Sequential sample generation of is exploited by \emph{adaptive IS} methods to adapt the proposal distribution based on previous samples, making the proposal closer to the target, improving sampling efficiency \cite{karamchandani89}.

Current state-of-the-art methods combine the previous ideas into what is know as \emph{adaptive multiple-IS} algorithms. Different sampling strategies, weighting schemes, and most importantly adaptation algorithms, result in several methods with different performance and properties \cite{LAIS, M-PMC, AMIS, APIS, PMC}. A recent review that includes the aforementioned methods can be found in \cite{Bugallo17}. We propose a new adaptation algorithm and a proposal distribution parameterized by a tree pyramid that in combination, improve the performance over previous methods.

\subsection{Stratified and quasi-Monte Carlo methods}
For low dimensional problems, there are deterministic procedures with better convergence rate and accuracy than MC and IS methods, examples are: \emph{generalized stratified sampling} \cite{wessing17}, \emph{latin hypercube sampling} \cite{Owen2013} and other \emph{grid based sampling} methods \cite{joshi16}. However, deterministic sampling strategies scale even worse than IS with dimensionality and do not adapt iteratively as new samples are generated. The algorithm proposed in this paper is at the intersection of \emph{stratified sampling} and \emph{Multi-Adaptive IS} by using a non-uniform multi-resolution tree structure to define the strata that parameterize the mixture components of the proposal distribution.

The usage of tree structures has been explored in the sampling literature, especially in the computer graphics application domain. Agarwal et. al. use a hierarchical representation of the image space to create different regular strata, the subdivisions are guided by a custom importance metric \cite{Agarwal03}. Clarberg et. al propose an approach based on wavelet function decomposition that is hierarchically constructed using a KD-tree. After its construction, it is used to warp an initial grid of samples towards regions with more light \cite{Clarberg05}. More recently, Estevez and Kulla have shown the computational benefits of a method with an adaptive tree strategy sampling that selects the light source used to compute a sample for a pixel value \cite{conty18}. Canevet et. al. inspired by Monte-Carlo Tree Search, proposed an adaptive binary tree to reduce training time in neural networks, the idea focuses on generating samples from the training dataset that are more important, weighting them using statistics from the loss function \cite{Canevet16}. These approaches are related with the method proposed in this paper, but are tailored for their application specific case.


\section{Adaptive importance sampling with tree pyramids}
\label{sec:methods}
As discussed in Section~\ref{sec:intro}, IS methods focus on approximating unknown values (of an arbitrary function $f(x)$) such as expected value, variance or skewness of an unknown PDF $\pi$. Those computations often involve intractable integrals that are approximated with a Monte Carlo estimator built sampling from $\pi$:
$$E_{\pi}[f(x)] = \int{f(x)\pi(x)dx} \stackrel{x \sim \mathcal{\pi}}{\approx} \frac{1}{N}\sum_{i=0}^{N-1} f(X_i).$$

When direct sampling from $\pi$ is not possible, IS draws samples $x$ from a known proposal distribution $x \sim \mathcal{Q}$ and weights them by the ratio of likelihoods $\pi(x)/\mathcal{Q}(x)$, namely the importance weight:

\begin{equation}
    E_{\pi}[f(x)] =  \int{f(x)\frac{\pi(x)}{\mathcal{Q}(x)}\mathcal{Q}(x)dx} \stackrel{x \sim \mathcal{Q}}{\approx} \frac{1}{N} \sum_{i=0}^{N-1} f(X_i)\frac{\pi(X_i)}{\mathcal{Q}(X_i)},
\end{equation}

where the variance of the IS estimator is determined by:
$$ Var(x) = \frac{1}{N}\sum_{i=0}^{N-1} \left[ f(X_i)\frac{\pi(X_i)}{\mathcal{Q}(X_i)} - E_\mathcal{Q}\right]^2.$$

Therefore, the selection of the proposal distribution $\mathcal{Q}$ has a huge impact on the sample efficiency by directly influencing the estimator variance. Usually the closer $\mathcal{Q}$ and $\pi$ are, the better the method performs. It is common for sampling methods to have an iterative nature (e.g. Monte Carlo Markov Chain). Adaptive methods, take advantage of previous samples drawn from $\mathcal{Q}_i$ to compute a new $\mathcal{Q}_{i+1}$ that is closer to $\pi$ to improve the estimator performance.

The sampling method proposed in this paper uses Tree Pyramids to describe the proposal distribution $\mathcal{Q}$ which focuses on generating samples in regions of high probability density in order to efficiently approximate the target posterior PDF $\pi$. After each sampling step, the tree is expanded to accommodate the new sample and improve the approximation of the proposal to the target distribution.

\subsection{Tree pyramid parameterized proposal distribution}
A K-dimensional tree-pyramid (KD-TP) is a full tree where each node $n=\{c_n,r_n, x_n, w_n, s_n\}$ represents a K-dimensional subspace with center $c_n$, radius $r_n$, sample $x_n$ with corresponding importance weight $w_n$, and a pointer $s_n$ to the set of children of $n$. $r_n$ is given by $r_n = r / 2^l$, where $r$ is the radius of the root node and $l$ is the node level. A node has either zero or $2^K$ children: $|s_n|= 0 \lor |s_n|=2^K $. The span of a tree is limited by the root radius. However, it is possible to dynamically re-root the tree to double the radius of the representable subspace in each dimension. The most common instances of KD-TP are Full Binary Trees($K=1$), Quadtrees ($K=2$) and Octrees ($K=3$).

\todo{OT: Feels like needs an introduction and a goal. NA:+1}
Without loss of generality, to simplify notation and implementation, we use the same radius for all dimensions making the subspaces hyper-cubes. This is not a limitation of the method as it is possible to represent the subspaces with different hyper-cuboids by allowing dimension-wise radii. If required, for example by the nature of the data, the implementation of dimension-wise radii should be a straightforward extension to the proposed method by defining $r \in \mathbb{R}^K$.

\begin{figure}
    \centering
    \includegraphics[width=0.9\columnwidth]{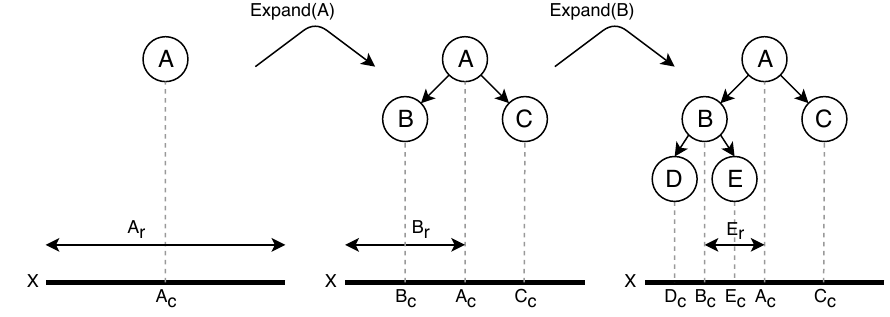}
    \caption{Example of construction of a 1D-TP over the domain X by leaf node expansion as implemented in Algorithm~\ref{alg:expansion}.}
    \label{fig:node_expansion}
\end{figure}

Creation and population of a KD-TP is described by the algorithm that expands a node, namely sub-divide one of the subspaces represented by a leaf node into its $2^K$ children, see Algorithm~\ref{alg:expansion}. An example of a 1-D tree construction by node expansion is shown in Figure~\ref{fig:node_expansion}.

The probability distribution parameterized by a KD-TP $\mathcal{T}$, is defined as a mixture model of distributions $D$:

\begin{equation}
\mathcal{Q}(x) = \sum^{|\lambda|-1}_{i=0} \bar{w}_iD(x;\lambda_{ic}, \lambda_{ir})
\label{eq:TPdist}
\end{equation}
\todo{NA: Add a superindex to denote the iteration. Consider adding other parameters to the equation and rethink the formulation. Nilesh will try to propose an alternative.}

where the number of mixture components $|\lambda|$ is determined by the cardinality of the set of leaf nodes $\lambda = \{n \in \mathcal{T} | n_s=\varnothing\}$ with the location of the kernels described by each leaf node center and the scale determined by its radius. For the experimental evaluation of the method, we plugged into Eq.~\ref{eq:TPdist} two types of distribution, a Uniform $D(\cdot)= U(\lambda_{ic}-\lambda_{ir}, \lambda_{ic}+\lambda_{ir})$ and a Multivariate Normal $D(\cdot)= \mathcal{N}(\lambda_{ic}, \lambda_{ir})$.

\begin{algorithm}[t]
    \caption{Leaf expansion algorithm for a K-Dimensional Tree Pyramid}
    \label{alg:TP_expand}
    \label{alg:expansion}
    \begin{algorithmic}[1]
        \Function{expandNode}{$n, K$}
        \State $\{\delta_{0}, \dots, \delta_{2^K-1}\} \leftarrow [\frac{r_n}{2},-\frac{r_n}{2}]^K$ \hfill\makebox[0.6\linewidth][l]{$\triangleright$ Cartesian product of $+r_n/2$ and $-r_n/2$ in $2^K$-dim.}
        \State $c_{child}^k \leftarrow c_n + \delta_{k}, k=1,\dots, 2^K-1$ \hfill\makebox[0.55\linewidth][l]{$\triangleright$ Obtain centers for the new $2^K$ nodes.}
        \State $r_{child} \leftarrow r_n/2$ \hfill\makebox[0.6\linewidth][l]{$\triangleright$ Compute the radii of the new nodes.}
        \State $n_s \leftarrow \{\{c_{child}^k,r_{child}\}\}_{k=1}^{2^K-1}$ \hfill\makebox[0.6\linewidth][l]{$\triangleright$ Assign new nodes as children of the expanding node.}
        \State \Return $n_s$ \hfill\makebox[0.6\linewidth][l]{$\triangleright$ Return the set of new nodes.}
        \EndFunction
    \end{algorithmic}
\end{algorithm}

\subsection{Quasi-Monte Carlo adaptive importance sampling with tree pyramids}

\begin{figure}
    \centering
    \includegraphics[width=0.99\textwidth]{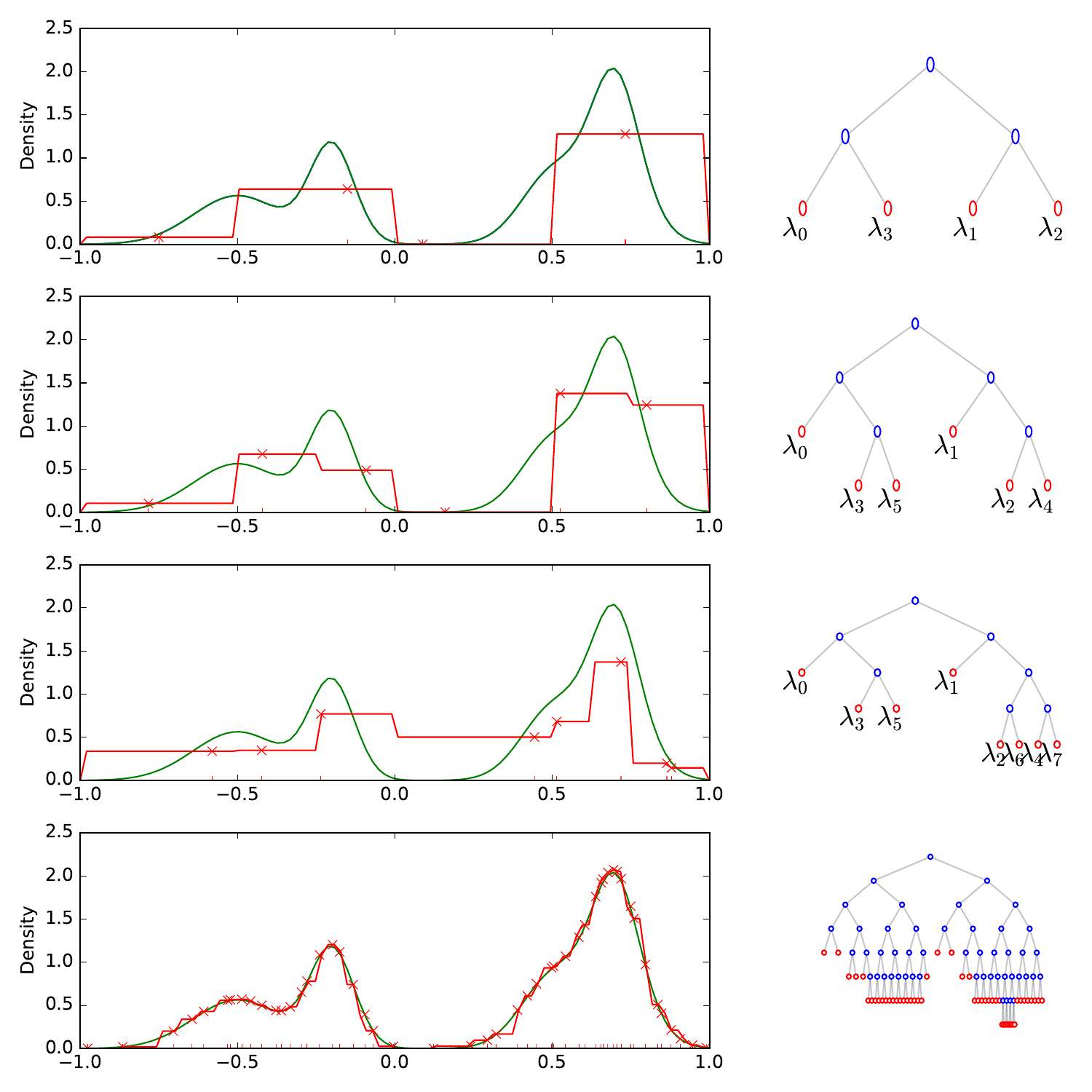}
    \caption{Construction of a 1D-TP. Left: Target distribution (green), proposal distribution (red) and samples from each leaf (red cross). Right: TP parameterizing the proposal distribution. Top to bottom: Sequence of node expansions following the procedure described in Algorithm~\ref{alg:TP} for N=4, N=6, N=8 and N=50.}
    \label{fig:sampling_example}
\end{figure}

Quasi-Monte Carlo methods lie between stratified and Monte Carlo sampling. In our method, the tree structure and the sub-division in subspaces is deterministic, while obtaining a sample from each hyper-cube is stochastic. This paper introduces the concept of non-uniform stratification that adapts the proposal distribution $\mathcal{Q}$ to the target distribution $\pi$. Figure~\ref{fig:sampling_example} shows a sampling and adaptation sequence. 

Like other AIS algorithms, TP-AIS can be divided in three major blocks: 1) Sampling, 2) weighting and 3) adaptation. Algorithm~\ref{alg:TP} describes the process of tree pyramid adaptive importance sampling that; 1) draws samples from the proposal distribution, parameterized by a tree pyramid, (line 15). 2) Computes the importance weights (line 16), and 3) adapts the proposal after new samples are drawn (line 18).

The resulting proposal distribution (in this case parameterized by the resulting tree $\mathcal{T}$ after $N$ samples) can be used as a generative model that approximates the target PDF. It is important to note that this sampling algorithm is anytime and there is no need to wait for $N$ samples to be generated to obtain a good approximation, the tree generation can be interrupted and the resulting set of samples $\mathcal{X}$, weights $\mathcal{W}$ and tree structure $\mathcal{T}$ used as an approximation to the posterior PDF. In what follows, TP-AIS (Algorithm~\ref{alg:TP}) is described in more detail.

First, the root node is created at the center of the space with a radius that spans the desired sampling space determined by the space limits $x_{min}$ and $x_{max}$ (lines 2-3). The sample representing the root subspace $x_{n_0}$, is drawn from the proposal (line 4). The importance weight $w_{n_0}$ of the initial sample is computed and the root node is updated with the sample and its weight (lines 5-6). The tree $\mathcal{T}$, the set of samples, $\mathcal{X}$, and weights $\mathcal{W}$ are initialized with the root node (lines 7-9).

Repeat the next steps until the desired number of samples $N$ are obtained: 1) Obtain the leaf node with maximum importance density, computed as the product of the volume of the leaf node hyper-cube and its importance weight, and expand it (lines 11-13). 2) For each new node, generate samples from the new mixture components (line 15), and compute their weights (line 16). The returned weights are self-normalized (line 21).

\begin{algorithm}[t]
	\caption{Tree pyramid adaptive importance sampling}
	\label{alg:TP}
	\begin{algorithmic}[1]
	    \Function{TreePyramidAIS}{$K, \mathcal{Q}, N, x_{min}, x_{max}$}
		\State $c_{n_0} \gets (x_{max}+x_{min})/2$ \hfill\makebox[0.5\linewidth][l] {$\triangleright$ Compute root node center and radius.}
		\State $r_{n_0} \leftarrow |x_{max}-x_{min}|/2$
		\State $x_{n_0} \gets x \sim \mathcal{Q}(c_{n_0},r_{n_0})$
		\State $w_{n_0} \leftarrow \pi(x_{n_0}) / \mathcal{Q}(x_{n_0}; c_{n_0},r_{n_0})$ \hfill\makebox[0.5\linewidth][l]{$\triangleright$ Compute the importance weight.}
		\State $n_0 \leftarrow \{c_{n_0},r_{n_0}, x_{n_0}, w_{n_0}, s_{n_0}=\varnothing\}$ \hfill\makebox[0.5\linewidth][l]{$\triangleright$ Define the root node.}
		\State $\mathcal{T} \leftarrow \{n_0\}$ \hfill\makebox[0.5\linewidth][l]{$\triangleright$ Initialize the tree with the root node.}
		\State $\mathcal{X} \gets \{x_{n_0}\}$ \hfill\makebox[0.5\linewidth][l]{$\triangleright$ Initialize the sample set from the root.}
		\State $\mathcal{W} \gets \{w_{n_0}\}$
		\While{$|\mathcal{X}| \le N$}
		    \State $\lambda \leftarrow \{n \in \mathcal{T} | s_n = \varnothing \}$ \hfill\makebox[0.5\linewidth][l]{$\triangleright$ Obtain the set of tree leaves.}
		    \State $\hat{n} \gets \arg\max_{n} (\pi(x_n)r_n^K,~ \forall n \in \lambda)$ \hfill\makebox[0.5\linewidth][l]{$\triangleright$ Choose node with max estimated evidence.}
    		\State $s_{\hat{n}} \leftarrow expandNode(\hat{n},K)$ \hfill\makebox[0.5\linewidth][l]{$\triangleright$ Expand into its children.}
    		\For{$n_s \in s_{\hat{n}}$}
        		\State $x_{n_s} \sim \mathcal{Q}(c_{n_s},r_{n_s})$ \hfill\makebox[0.5\linewidth][l]{$\triangleright$ Sample the proposal of each new leaf.}
        		\State $w_{n_s} \gets \frac{\pi(x_{n_s})} {\mathcal{Q}(x_{n_s}; c_{n_s},r_{n_s})}$ \hfill\makebox[0.5\linewidth][l]{$\triangleright$ Compute the importance weight.}
        		\State $s_{n_s} \gets \varnothing$
        		\State $\mathcal{T} \leftarrow \mathcal{T} \cup n_s$  \hfill\makebox[0.5\linewidth][l]{$\triangleright$ Insert expanded node into the Tree.}
        		\State $\mathcal{X} \leftarrow \mathcal{X} \cup x_{n_s}$ \hfill\makebox[0.5\linewidth][l]{$\triangleright$ Insert sample into the sample set.}
        		\State $\mathcal{W} \leftarrow \mathcal{W} \cup w_{n_s}$ 
    		\EndFor
		\EndWhile
		\State $\bar{\mathcal{W}} \leftarrow \mathcal{W}_i / \sum_{i=0}^{|\mathcal{X}|-1}\mathcal{W}_i$ \hfill\makebox[0.5\linewidth][l]{$\triangleright$ Compute the self-normalized importance weights.}
		\State \Return $\mathcal{X}, \bar{\mathcal{W}}$
		\EndFunction
	\end{algorithmic}
\end{algorithm}

\subsubsection{Weighting schemes}
\todo{OT: Do we quantify the different flavors of the algorithm anywhere? JFL: Not yet, I think the implementation of the two flavors has a bug and the results are not convincing, I'll see if i can fix it and comment on the results.}

Depending on the choice for sampling, weighting and resampling, different flavors of TP-AIS can be implemented. The algorithm described in Algorithm~\ref{alg:TP}, corresponds to the \emph{standard} TP-AIS (sTP-AIS) which is an implementation that follows a weighting scheme known as \emph{standard} in multiple importance sampling \cite{Cappe04}, where importance weights are computed using each individual mixture component as the proposal distribution,

$$w_i = \frac{\pi(x_i)}{\mathcal{Q}_i(x_i)},$$

and self-normalizing them:

$$\bar{w}_i = \frac{w_i}{\sum_{i=0}^{N-1}w_i}.$$

Alternatively, weights can be computed using all the mixture components:

\begin{equation}
\label{eq:DMweights}
w_i = \frac{\pi(x_i)}{\frac{1}{N}\sum_{q=0}^{N-1}\mathcal{Q}_q(x_i)},    
\end{equation}

obtaining the Deterministic Mixture (DM) weighting. DM weights come at a more expensive computational cost (require the evaluation of all the proposals to compute weights) but are known to perform better in terms of variance \cite{elvira19}. DM-TP-AIS can be implemented by plugging Equation~\ref{eq:DMweights} into lines 5 and 16 of Algorithm~\ref{alg:TP}. In the case of choosing a Uniform distribution to represent each of the subspace proposal distributions, the DM approach reduces to the simple case after normalizing the importance weights.

\subsubsection{Leaf resampling}
Another variation to the sTP-AIS is the resampling strategy. As described in Algorithm~\ref{alg:TP}, each time a sample is generated in a new subspace, it is never removed from the sample set $\chi$ or replaced by another sample. This can be problematic if the target density $\pi$ within a subspace is peaked. A sample in such a subspace is likely to end up in a low density region. Consequently, the evidence in that space will be underestimated (line 12, Algorithm~\ref{alg:TP}), which may result in that subspace being selected for expansion much later than it should have.

This problem can be mitigated by implementing a resampling strategy. Before obtaining a new set of tree leaves (line 11, Algorithm~\ref{alg:TP}), all the leaf nodes of the tree are resampled and weighted. For each leaf node $n_s$, the new sample is retained only if its importance weight is higher than the existing importance weight $w_{n_s}$. This leads to a lower acceptance rate but reduces the variance, and increases robustness to multi-modal target distributions and subspaces represented by samples in low probability regions.

\subsubsection{Mixture TP-AIS}
Instead of ordering the tree leaves by importance density to select which one to expand, it is possible to generate the new samples directly from the mixture distribution parameterized by the tree pyramid. This sampling approach is similar to the Mixture Population Monte Carlo (M-PMC \cite{M-PMC}) approach, which, instead of generating one new sample from each individual component of the proposal, samples from the mixture itself. To implement this approach, the selection of the node to expand depends on a sample drawn from the weighted mixture model. The mixture weights depend on the normalized importance volume of the component:
$$ w_i = \frac{\lambda_{iw}\lambda_{ir}^K}{\sum_{j=0}^{N-1}\lambda_{jw}\lambda_{jr}^K}.$$

\todo{Comment the diffrences and a plot showing the different flavors. sTP-AIS(r), DM-TP-AIS(r), M-TP-AIS(r)}

Then, a sample from the weighted mixture model is obtained by Algorithm~\ref{alg:sampleMixture} in a two-step process. First, a random number $\alpha \sim U(0,1)$ is generated to select the sampling distribution (lines 3-6). Second, the sample is obtained by sampling from the selected $D$ mixture component. This approach replaces the sort operation in line 9 of Algorithm~\ref{alg:TP} by a sampling operation and adds a tree search operation to select which node to expand.

\begin{algorithm}
	\caption{Weighted mixture model sampling.}
	\label{alg:sampleMixture}
	\begin{algorithmic}[1]
		\Function{SampleMixtureModel}{$D, \lambda, N$}
		\State  $w_i = \lambda_{iw}\lambda_{ir}^K / \sum_{j=0}^{N-1}\lambda_{jw}\lambda_{jr}^K$
		\State  $\alpha \sim U(0,1)$
		\State  $N \leftarrow 0$
		\While{$\alpha < \sum_{i=0}^N w_i$}
    		\State  $N \leftarrow N + 1$
		\EndWhile
		\State \Return $x \sim D_N$
		\EndFunction
	\end{algorithmic}
\end{algorithm}

\section{Evaluation}
\label{sec:evaluation}

\subsection{Baselines}
\todo{OT: We should summarize results somewhere here.}
We compare our TP-AIS method introduced in Section~\ref{sec:methods} to a variety of state-of-the-art AIS methods: LAIS \cite{LAIS}, APIS \cite{APIS}, M-PMC \cite{M-PMC}, DM-PMC \cite{DM-PMC}. The comparison is extended beyond AIS methods and includes an MCMC baseline with the Metropolis-Hastings algorithm \cite{Hastings70} and Multi-Nested sampling \cite{Feroz2014} which is an evolution of Nested sampling \cite{Skilling2006}.

\subsection{Evaluation metrics}
When the task at hand is to draw samples from a target distribution, the Effective Sample Size (ESS) is a popular measure of sampling efficiency for MCMC and IS methods. By normalizing it by the number of total samples N, we obtain the Normalized Effective Sample Size (N-ESS) which can be interpreted as a factor that determines the number of samples required from the evaluated algorithm to generate an independent sample from the target distribution. This metric takes into account that samples in IS are biased by the proposal distribution, and MCMC samples are autocorrelated. In IS, an approximation $\widehat{ESS}$ is often defined as the inverse sum of the squared normalized importance weights:

\todo{NA: Mention the acceptance rate and whether to use it or not. We want to highlight that the ESS is good despite the not so low acceptance rate. Comparing the method with rejection sampling will be helpful.}

$$\widehat{ESS} = \frac{1}{\sum_{i=0}^{N-1}\bar{w}^2_i},$$

in MCMC the definition of ESS is:

$$ESS = \frac{N}{1+2\sum_{i=0}^{N-1}\rho(i)}$$

Where $N$ is the number of samples and $\rho(i)$ is the correlation at the i-th MCMC step. Although this metric has received some criticism, it is still widely used and it generally provides good performance \cite{Martino17}. To obtain the N-ESS we just divide the $ESS$ or the $\widehat{ESS}$ by the total number of samples $N$.

Besides ESS, we evaluate the approximation quality of the adapted proposal distributions obtained by AIS methods with a probability distribution similarity metric like the Jensen Shannon Divergence (a symmetric version of the more popular Kullback–Leibler divergence) shown in Eq.~\ref{eq:cont_JS}.

\begin{equation}
    D_{KL}(P || Q) = \int p(x)\log \left( \frac{p(x)}{q(x)}\right) dx
    \label{eq:cont_KL}
\end{equation}

\begin{equation}
    D_{JS}(P || Q) = \frac{1}{2} D_{KL}\left(P || \frac{P+Q}{2}\right) + \frac{1}{2} D_{KL}\left(Q || \frac{P+Q}{2}\right)
    \label{eq:cont_JS}
\end{equation}

Unfortunately, both similarity metrics require the computation of integrals of intractable densities. Because the parameter space we are considering is bounded, we could approximate them with enough precision by considering a small value of $dx$. However, it becomes rapidly intractable when the number of dimensions increases. To address this issue, we use a Monte Carlo estimate to the metrics by drawing $N$ uniformly distributed samples from P and Q and compute their empirical distance metrics, See Eq.\ref{eq:MC_KL}.
\begin{equation}
    \hat{D}_{KL}(P || Q) = \frac{1}{N}\sum^{N-1}_{i=0} p(x_i)\log \left( \frac{p(x_i)}{q(x_i)}\right),
    \label{eq:MC_KL}
\end{equation}

\begin{figure}
    \centering
    \includegraphics[width=0.99\columnwidth]{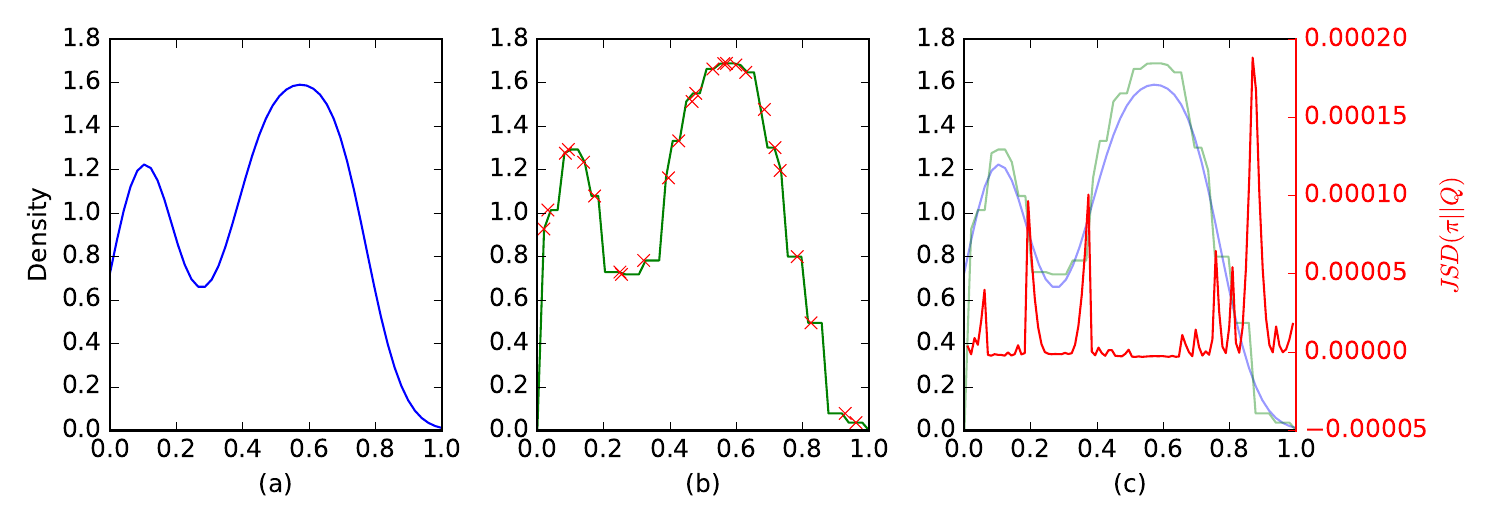}
    \vspace{-0.4cm}
    \caption{Example evaluation of our sTP-AIS with N=25. a) Ground truth PDF $\pi$. b) Importance samples (red) generated from TP sampling algorithm and proposal PDF $\mathcal{Q}$ (green). c) Overlaid ground truth distribution (blue) and adapted proposal  (green) with the corresponding Jensen-Shannon Divergence (red).}
    \label{fig:sampling_evaluation}
\end{figure}

An example evaluation with N=25 is depicted in Figure~\ref{fig:sampling_evaluation}. The evaluation proceeds as follows:
\todo{OT: Can we make a case for acceleration based on this? Not on the paper, this is a comment to keep in mind.}

\begin{enumerate}[(i)]
    \item Select a known PDF $\pi$ as the ground truth distribution.
    \item Use the evaluated sampling method $M$ to draw $N$ samples.
    \item If the IS method used is adaptive, the proposal distribution $\mathcal{Q}$ after drawing $N$ samples is used. If the evaluated method only generates samples and does not adapt the proposal, the drawn samples are used to compute a KDE approximate PDF $\mathcal{Q}^*$.
    \item Compute evaluation metrics: ESS($M$), elapsed time and $JSD(\pi||\mathcal{Q})$.
\end{enumerate}

The MCMC based baseline methods do not have a proposal distribution that is being adapted to approximate the target distribution, in order to evaluate how close the samples generated are to the target distribution, we use a Kernel Density Estimate (KDE) using the generated samples.

Some sampling methods are based on accept/reject \cite{Hastings70, Skilling2006} or have complex sampling mechanisms \cite{Feroz2014} that can yield good ESS with an increased computational cost, hence we include runtime to the evaluation metrics. This is of special interest in cases where the likelihood function computational cost is negligible and the sampling strategy becomes the computational bottleneck. 

\subsection{Ground truth distributions}
\begin{figure}
    \centering
    \includegraphics[width=0.99\textwidth]{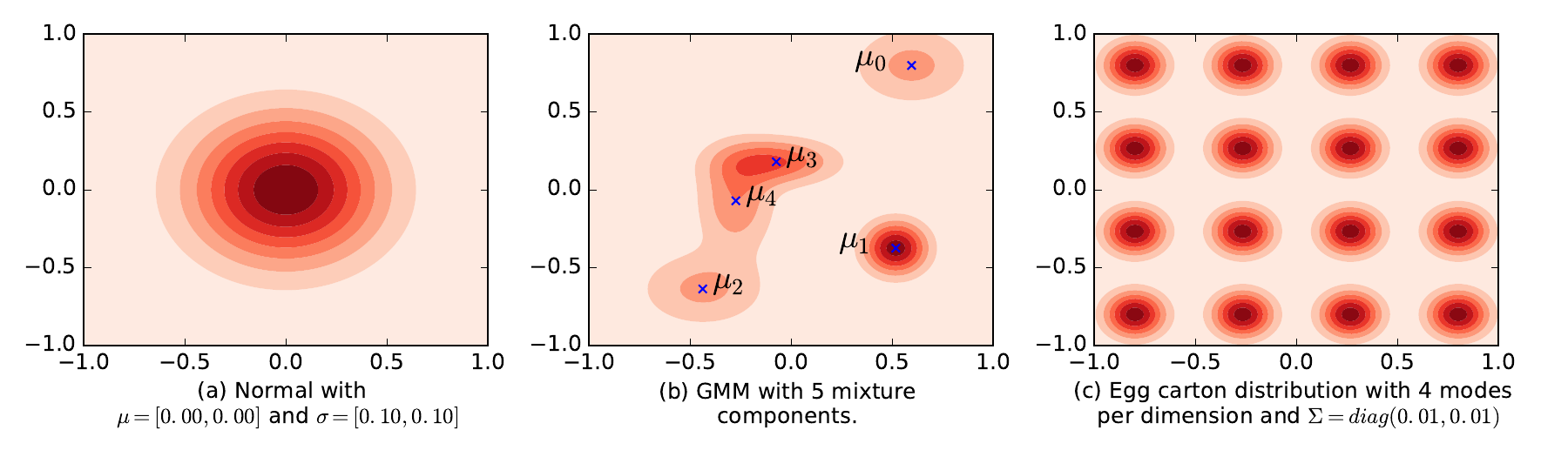}
    \caption{Example of 2D ground truth PDFs used to evaluate the sampling methods.}
    \label{fig:target_pdfs}
\end{figure}

Sampling methods performance often depends on the properties of the target distribution $\pi$. Thus, we evaluate the methods using different ground truth parametric PDFs with varied number of modes and randomized moments, see some examples in Figure~\ref{fig:target_pdfs}. We use Gaussian Mixture Models (GMMs) to parameterize the ground truth distributions as follows:

    $$\pi_\mathcal{N}(x|\mu, \Sigma) = \frac{1}{\sqrt{(2\pi)^n |\Sigma|}} exp\left(-\frac{1}{2}(x-\mu)^T \Sigma^{-1}(x-\mu)\right),$$

    $$\pi_{gmm}(x|\mu, \Sigma, \bar{\omega}) = \sum^{n-1}_{i=0} \bar{\omega}_i \pi_\mathcal{N}(x|\mu_i, \Sigma_i).$$

Where $\mu_i \in \mathbb{R}^d$ is a vector of means, $\Sigma_i \in \mathbb{R}^{d \times d}$ is a covariance matrix that parameterize a gaussian kernel $\pi_\mathcal{N}$. $\bar{\omega}_i \in \mathbb{R}$ is the normalized mixture weight. The sub-index $i$ is used to reference the $i^{th}$ mixture component of the $n$ components of dimensionality $d$ that compose the mixture model. Examples of 2-D GMMs used for evaluation are shown in Figure~\ref{fig:target_pdfs}.

For the evaluation, three different types of GMMs with randomized first and second order moments are used: i) The first target distribution, referred as "normal", has just one mixture component with $\mu \sim U(-1,1)$ and $\sigma \sim U(0.01,0.05)$. ii) the second target distribution has 5 mixture components with $\mu \sim U(-1,1)$ and $\Sigma \sim diag( U(0.01,0.05) )$. iii) the third target distribution is a GMM version of the egg crate function with 4 equidistant modes per dimension and $\Sigma = diag(0.01)$. An example of each ground truth distribution is depicted in Figure~\ref{fig:target_pdfs}.

\begin{figure}[t]
    \centering
    \includegraphics[width=0.99\columnwidth]{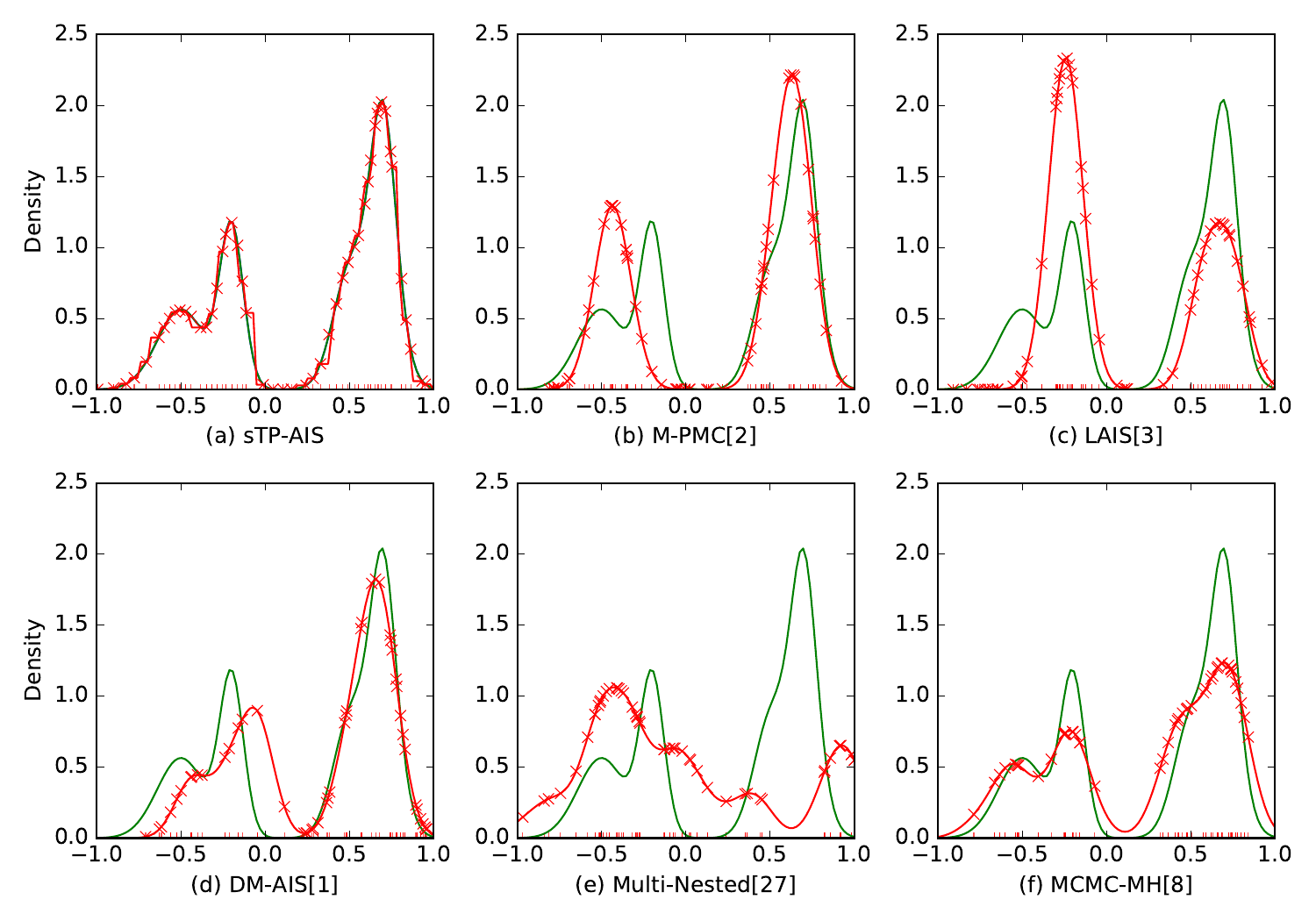}
    \caption{Qualitative comparison of our TP-AIS and the baseline methods after 50 samples for a Gaussian Mixture Model target distribution (shown in green). The proposal distribution after 50 samples in shown in red, with the 50 samples drawn as red crosses.}
    \label{fig:qualitative}
\end{figure}


\section{Results and discussion}
\label{sec:results}
\begin{figure}[t]
    \centering
    \includegraphics[width=0.49\columnwidth]{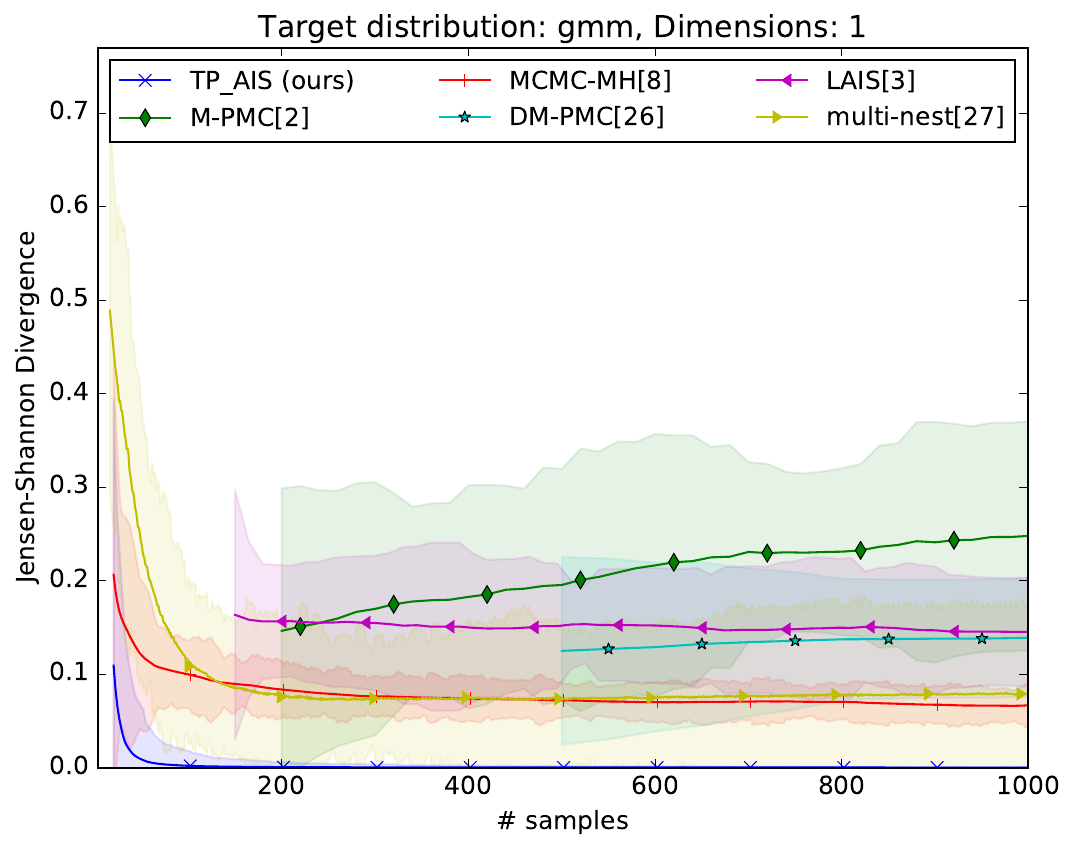}
    \includegraphics[width=0.49\columnwidth]{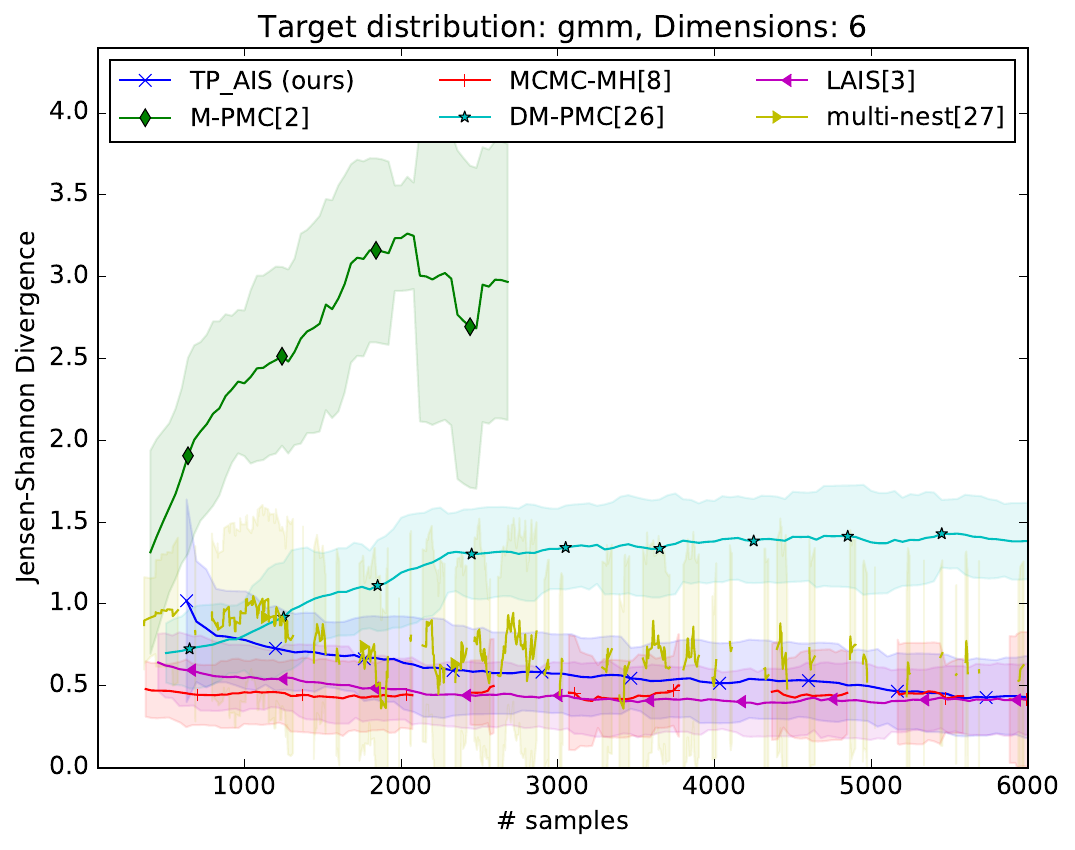}
    \includegraphics[width=0.49\columnwidth]{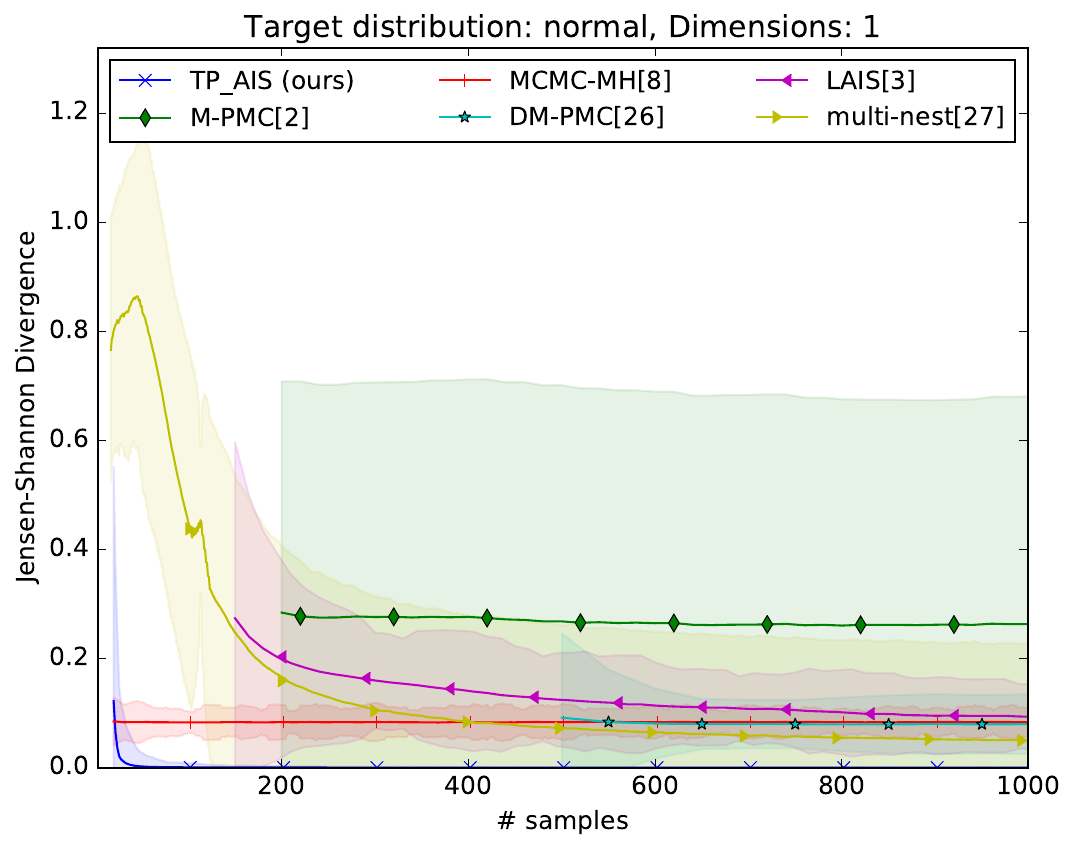}
    \includegraphics[width=0.49\columnwidth]{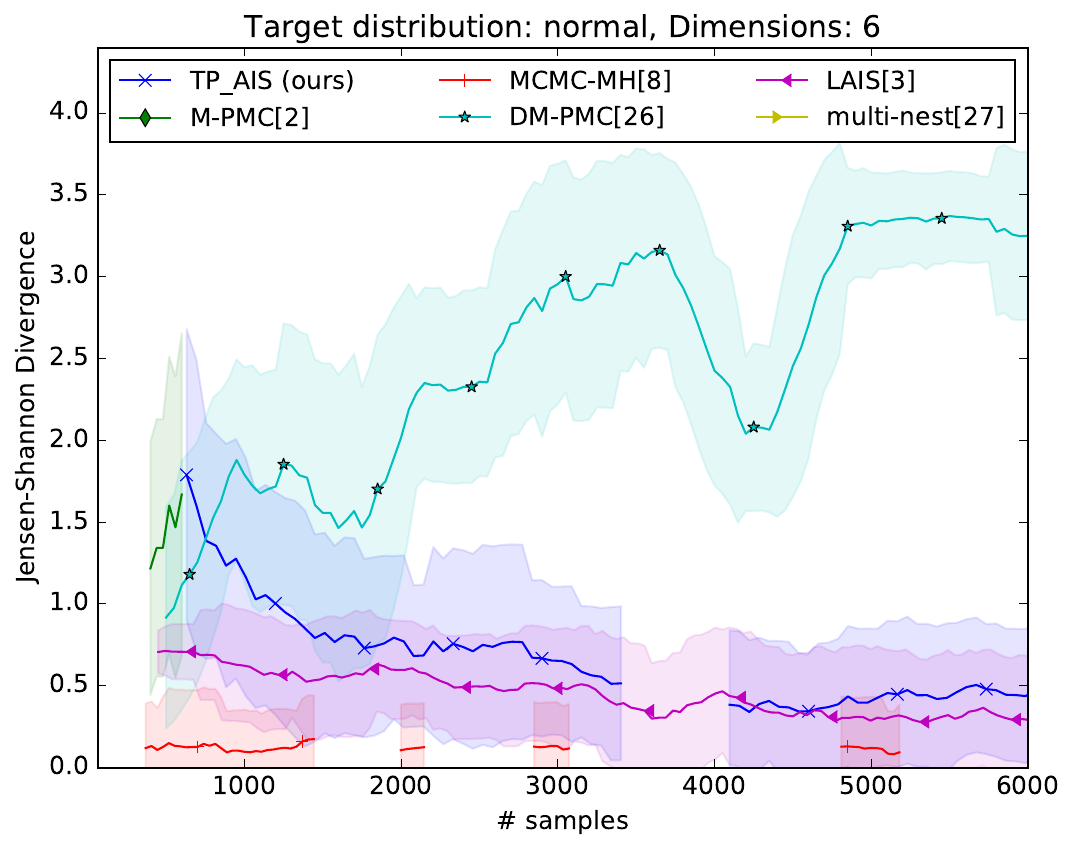}
    \caption{Experimental result statistics for the Jensen-Shannon Divergence metric for different target distributions and dimensionality.}
    \label{fig:resultsJSD}
\end{figure}

\begin{figure}[t]
    \centering
    \includegraphics[width=0.49\columnwidth]{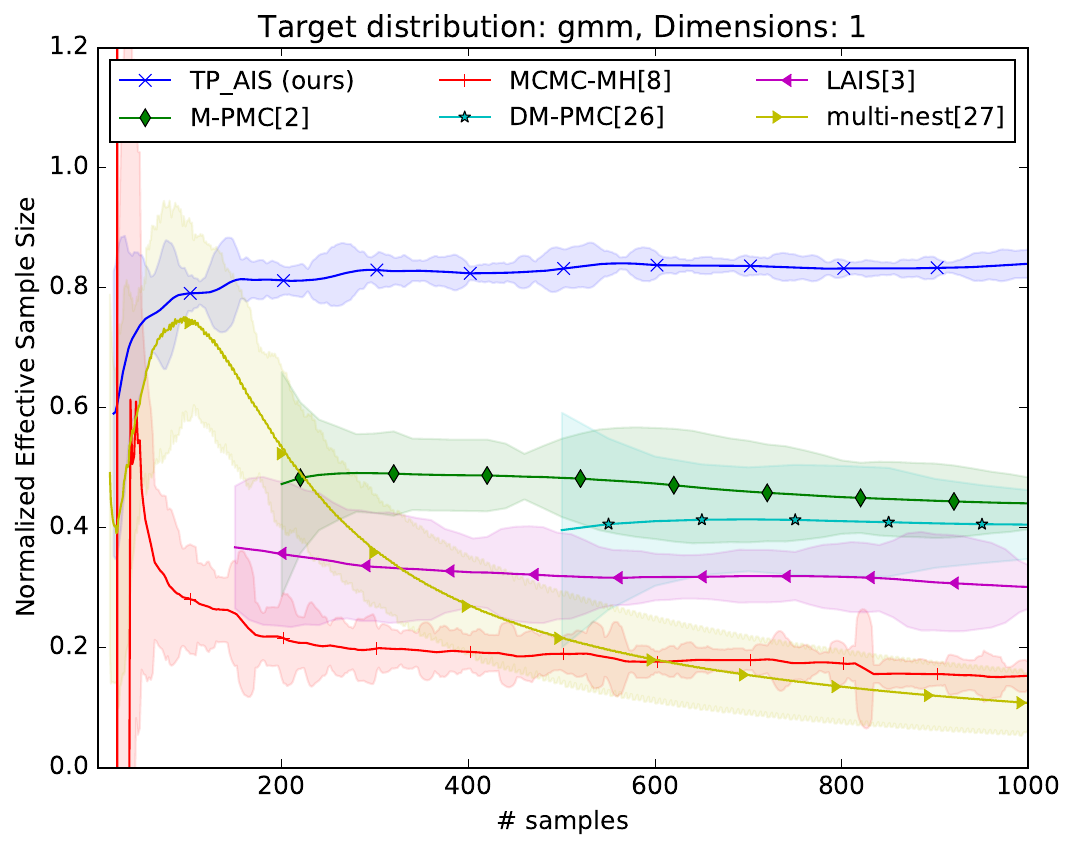}
    \includegraphics[width=0.49\columnwidth]{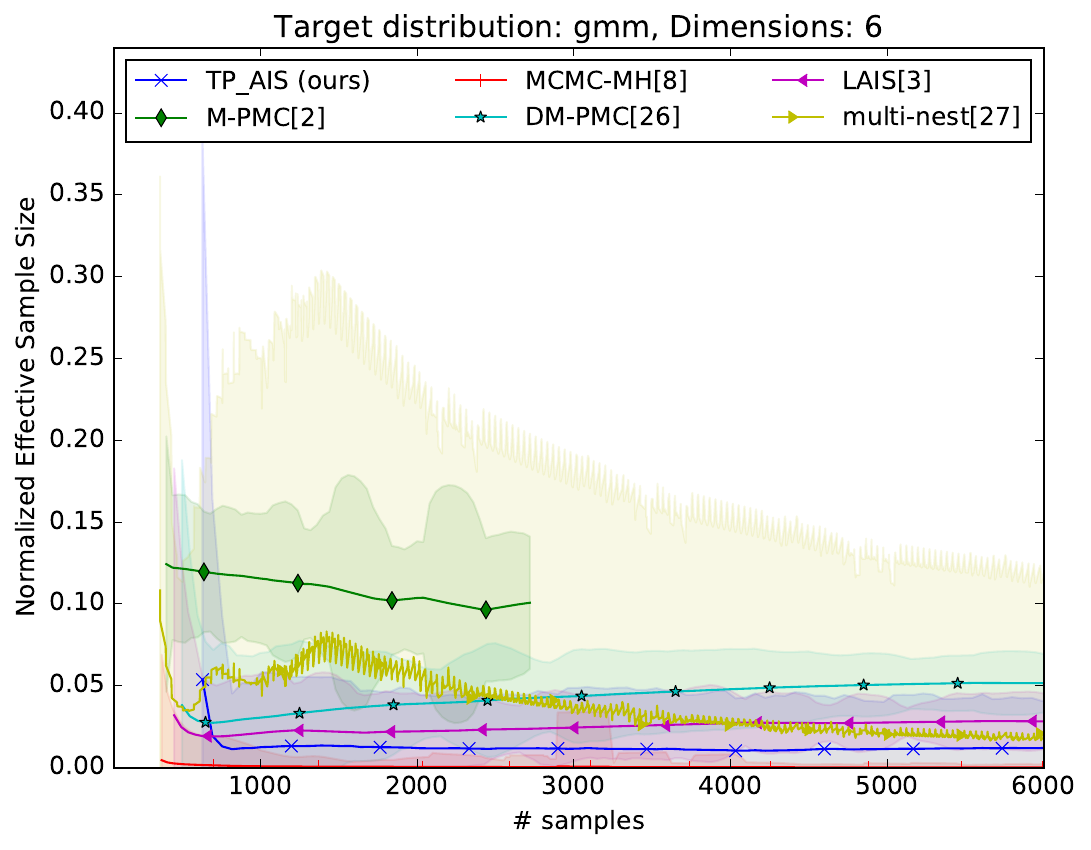}
    \includegraphics[width=0.49\columnwidth]{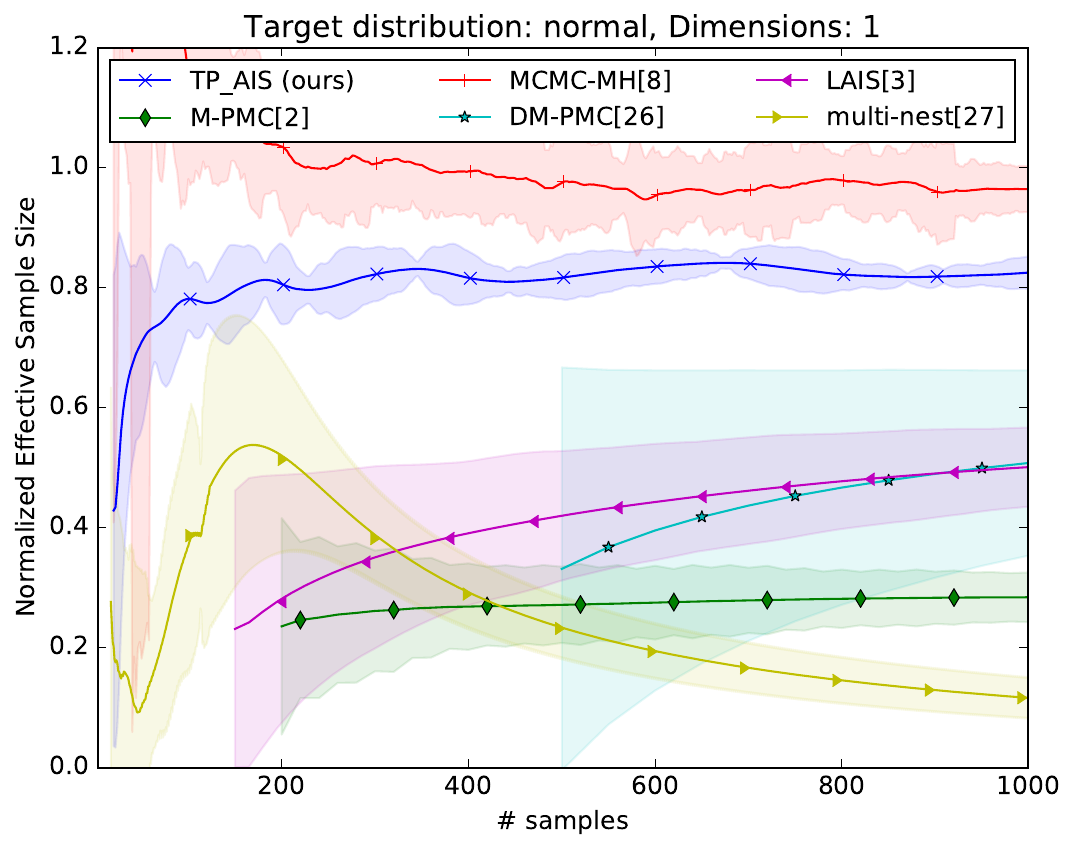}
    \includegraphics[width=0.49\columnwidth]{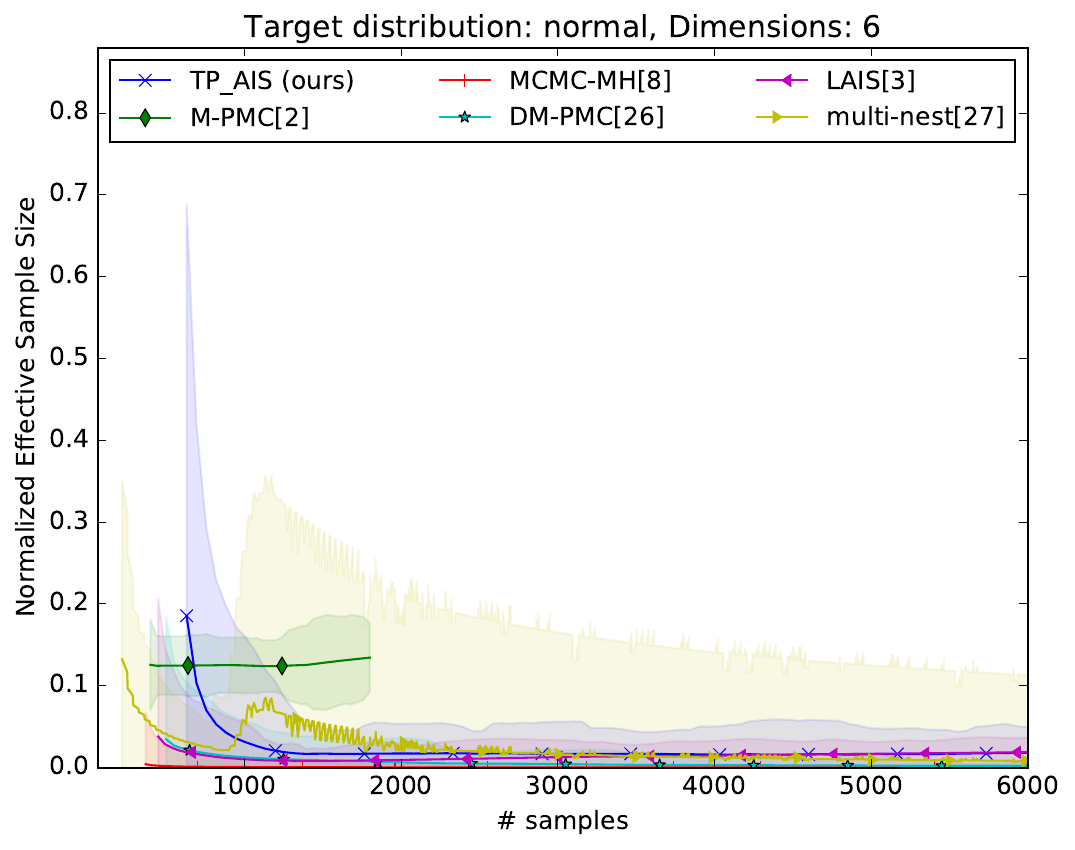}
    \caption{Experimental result statistics for the Normalized Effective Sample Size metric for different target distributions and dimensionality.}
    \label{fig:resultsNESS}
\end{figure}

\begin{figure}[t]
    \centering
    \includegraphics[width=0.49\columnwidth]{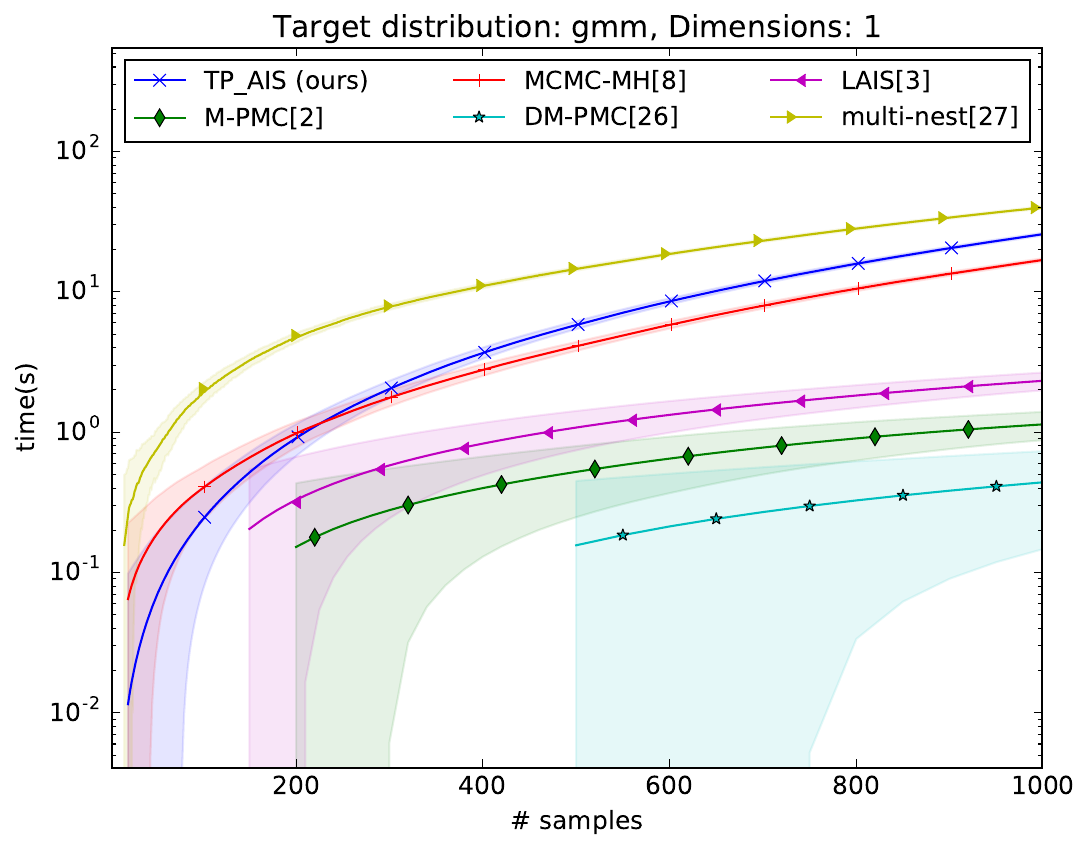}
    \includegraphics[width=0.49\columnwidth]{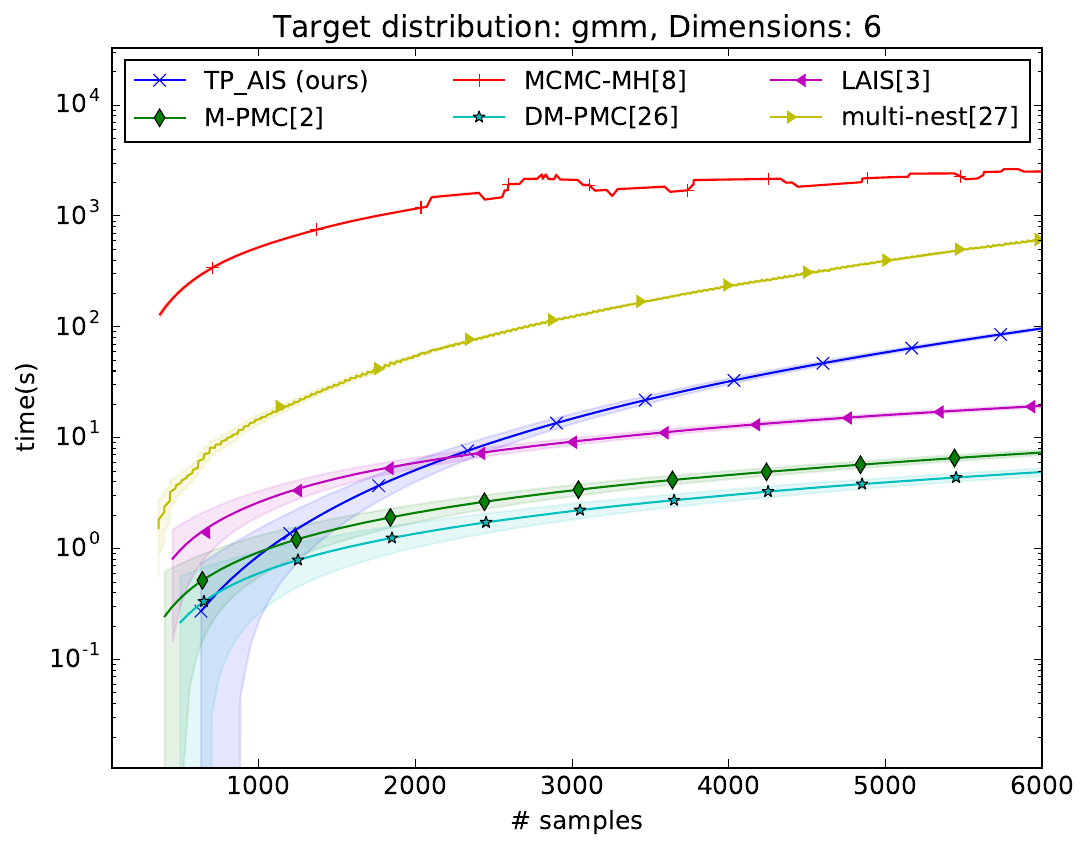}
    \includegraphics[width=0.49\columnwidth]{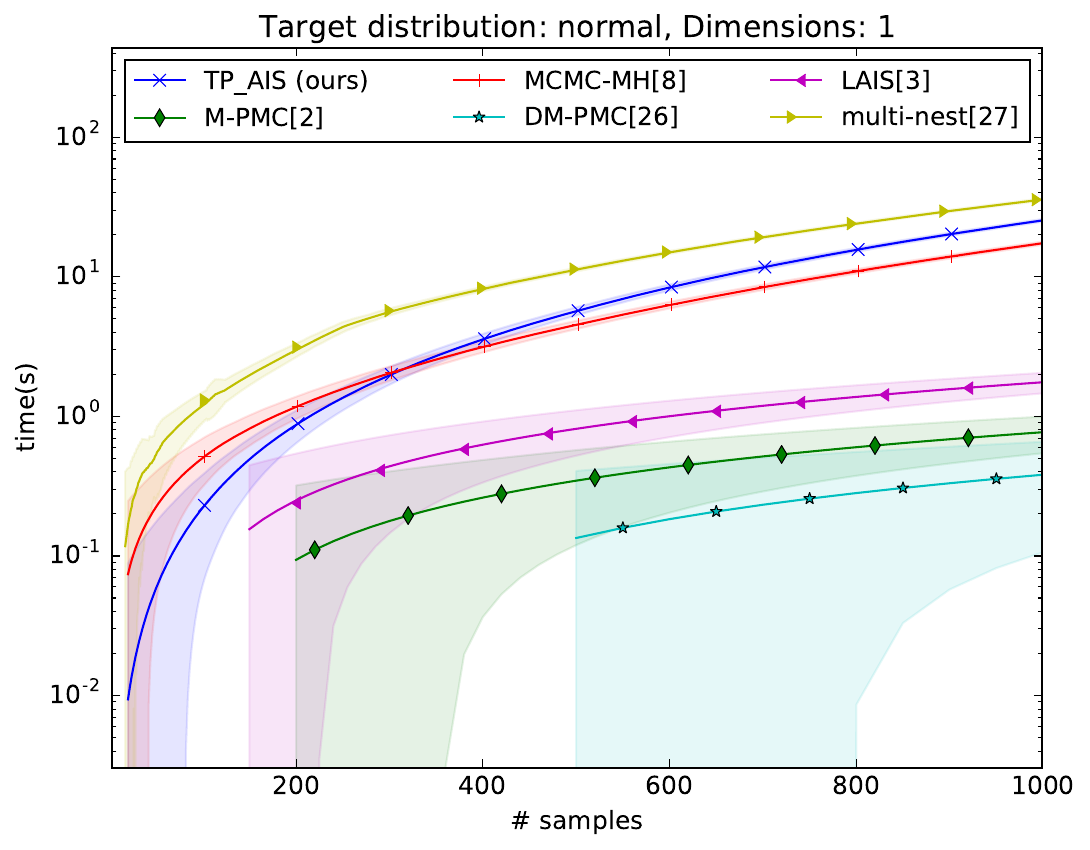}
    \includegraphics[width=0.49\columnwidth]{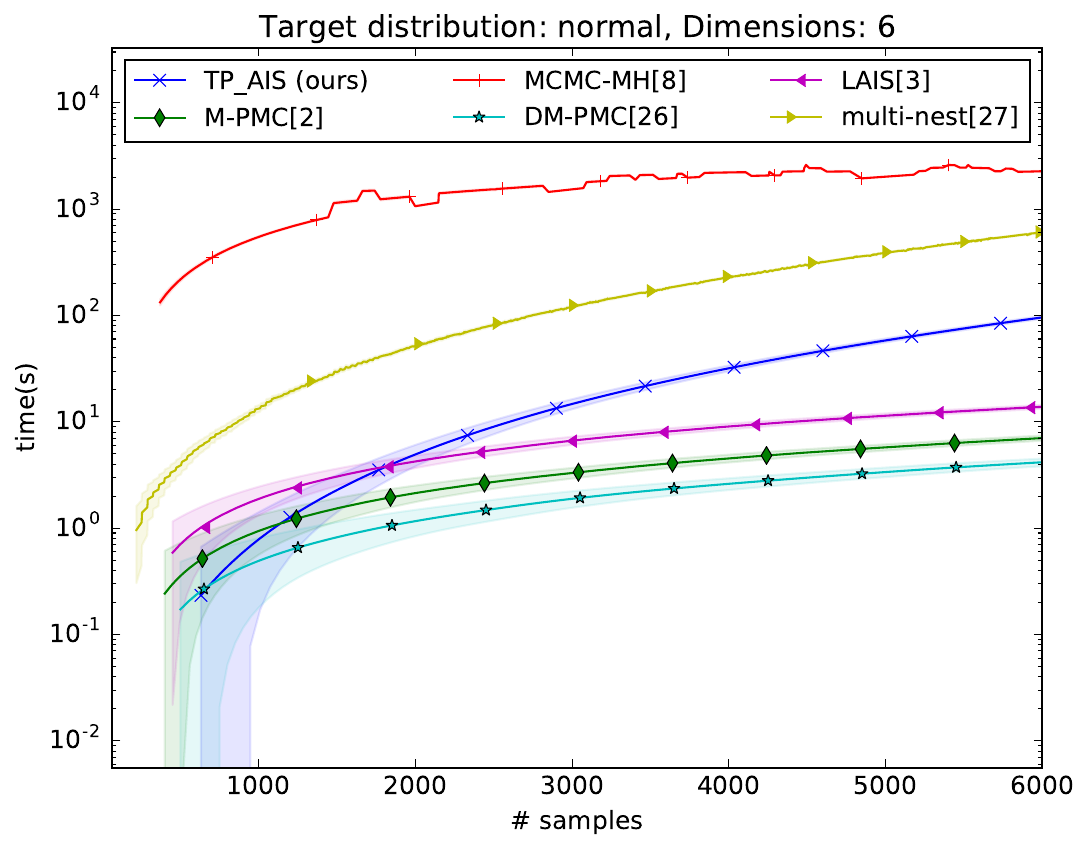}
    \caption{Experimental result statistics for the computational time complexity metric for different target distributions and dimensionality.}
    \label{fig:resultsTime}
\end{figure}

To obtain the results presented in this section, we have conducted the procedure detailed in Section~\ref{sec:evaluation} a hundred times for each combination of dimensionality and random sampled target PDF. The random seed was fixed to evaluate all the methods with the same target distributions. The results (Figures~\ref{fig:resultsJSD}-\ref{fig:resultsTime}) show the empirical mean as a solid line and the stdev as a shaded color. Each experiment was executed with a time limit. In some cases, the evaluated methods were unable to provide the required number of samples. As a consequence there are plots that lack data points, which can be considered failure cases that we plan to analyze as part of our next steps. We have considered the simple version of TP-AIS with resampling which will be the flavor of TP-AIS being used unless otherwise mentioned. For more results with other variations of the method we refer the reader to the additional material.

In Figure~\ref{fig:qualitative} we show a visualization of the posterior approximation obtained with the different methods evaluated. It can be seen how the TP-AIS approach (Figure~\ref{fig:qualitative}(a)) provides a closer approximation to the target distribution than the other methods with the same number of samples (Figure~\ref{fig:qualitative}(b-f)). In the remainder of this section the evaluated metrics vs. number of samples are discussed.

One of the most impactful aspects on performance in importance sampling is the similarity between the proposal and target distributions. We have evaluated the similarity using the Jensen-Shannon Divergence detailed in Section~\ref{sec:evaluation}. For lower dimensional problems, the left column of Figure~\ref{fig:resultsJSD} shows that our TP-AIS method outperforms the baselines by converging much faster to a better approximate solution in the three types of distributions evaluated. For higher dimensional cases, TP-AIS performance is on par with other methods, see right column of Figure~\ref{fig:resultsJSD}.

\label{ssec:resultsJSD}

N-ESS measures the sample efficiency of a method, results for this metric are shown in Figure~\ref{fig:resultsNESS}, TP-AIS outperforms existing methods in terms of sample efficiency in low dimensions. In higher dimensions all the methods yield poor performance as can be seen in the right column of Figure~\ref{fig:resultsNESS}. This low performance is an expected result, for higher dimensions density functions become more concentrated causing the importance weights to sky-rocket, consequently reducing the ESS.

\label{ssec:resultsNESS}

Regarding the time complexity metric, Figure~\ref{fig:resultsTime} shows how the evaluated methods scale with the number of samples. Although TP-AIS implements a more sophisticated sampling method, the time required to generate N samples is comparable to the rest of the methods for a reasonable number of samples. A surprising result observed in higher dimensional problems (Figure~\ref{fig:resultsTime} right), is the sub-par performance of the MCMC-MH approach, where it is supposed to scale well with dimensions. With the increase in dimensions, the target density is more concentrated causing a lot of the proposed MC moves to have very low acceptance probability and be rejected, dramatically increasing the time required to generate N samples. This is a well known problem that MCMC methods suffer from and can be alleviated by the fine-tuning the MC proposal distribution. It can be seen how the other methods are not impacted by this issue.

\subsection*{Discussion}
The presented method provides a significant improvement in the sampling efficiency (See. Figure~\ref{fig:resultsNESS}). It might seem that the more complex formulation of the proposal distribution and its adaptation can impact the computational complexity of the sampling algorithm. However, although in some cases TP-AIS requires more time to generate a specific number of samples (especially when the desired number of samples is high), its sampling efficiency allows the method to obtain a better proposal distribution approximation with less samples. For example, in the 1D GMM case, shown in Figure~\ref{fig:resultsJSD}, TP-AIS obtains a better representation of the target $\pi$ (i.e. lower JSD) with 100 samples than the other benchmarked methods do with 1000 samples.

In contrast to other Quasi Monte-Carlo approaches \cite{wessing17, joshi16}, our proposed algorithm does not lose the anytime property that MCMC methods have, and the sampling process can be halted at anytime allowing for accuracy-time trade-offs.

The main drawback of the proposed approach is dimensionality, which limits its application from low to mid dimensional problems. Each new sample step subdivides the space into $2^K$ subspaces, constraining the number of samples that can be generated after each time step. For example, for a 9D problem, each sampling step needs to generate 512 samples which, depending on the application, might be overwhelming. On the other hand, this fact opens the door to parallel implementations that compute the likelihood of all the new samples concurrently further improving the number of samples per unit of time that this approach can deliver. In any case, this is a well-known limitation of IS methods in general and TP-AIS is not an exception.

Another limitation of TP-AIS is the need of boundaries for the sampling space. This can cause to have regions of non-zero probability outside the approximated space, the probability density outside the space is distributed over the sampling space resulting in density over-estimation. However, in practice it is reasonable to assume known boundaries in the domain of the sampled function and in our evaluation we did not experience this issue to be a limitation.

Besides defining the sampling domain, the parameter free aspect of TP-AIS is one of its strengths. All other methods have parameters that need to be tuned. It is well-known that the proposal distribution design has a huge impact on performance, because sampling algorithms parameters have an important impact on the adaptation of the proposal distribution, a good parameter selection has a huge influence on performance. TP-AIS circumvents that problem enabling to exploit its full potential without deep domain knowledge.


\section{Conclusions}
\label{sec:conclusion}
In this paper we have presented TP-AIS, a novel adaptive importance sampling algorithm that parameterizes the proposal distribution using tree pyramids which structure the proposal in partitioned subspaces. The evaluation with a variety of target posteriors shows its benefits in accuracy and sample efficiency over state-of-the-art and well-known approaches.

These are the first steps of a promising sampling scheme that combines quasi-Monte Carlo techniques with adaptive importance sampling. More research in the resampling strategies, kernel selection and subspace representation can lead to improved sample efficiency and approximations.

The convergence speed of this sampling technique and its high ESS may enable the application of IS for efficient inference in applications with mid-to-low dimensionality like object pose estimation and protein-ligand docking.

\bibliographystyle{unsrt}
\bibliography{references}

\clearpage
\section*{Additional material}
\section*{Additional material: Extended state-of-the-art comparison results}
\subsection*{Evaluated algorithm parameters}
In the evaluation section we compare our method with a set of classical and state-of-the-art methods. However, such methods need to be configured with multiple parameters in order to achieve reasonable performance. Because the evaluation process uses multiple target distributions with randomized moments, we have tuned the parameters of the benchmarked algorithms to achieve good performance. The parameters used with a brief description can be found below. For more details of the functionailty of each parameter we refer the reader to each of the papers describing the methods.

\begin{verbatim}
    M-PMC [2]
    K: 20         # Number of samples per proposal distribution
    N: 10         # Number of proposal distributions
    J: 1000       # Limit number of samples
    sigma: 0.01   # Scaling of the spherical gaussian proposal

    MCMC-MH [8]
    n_burnin: 10     # Number of burn-in samples
    n_steps: 2       # Number of samples to skip (a.k.a. decorrelation samples)
    prop_sigma: 0.01 # Scaling of the spherical gaussian transition
    kde_bw: 0.01     # KDE bandwidth for posterior approximation

    LAIS [3]
    K: 3          # Number of samples per proposal distribution
    N: 5          # Number of proposal distributions
    J: 1000       # Limit number of samples
    L: 10         # Number of MCMC moves during the proposal adaptation
    sigma: 0.01   # Scaling of the spherical gaussian proposal distributions
    mh_sigma: 0.005  # Scaling of the mcmc proposal distributions moment update

    DM-PMC [26]
    K: 10         # Number of samples per proposal distribution
    N: 2          # Number of proposal distributions
    J: 1000       # Limit number of samples
    sigma: 0.01   # Scaling of the spherical gaussian proposal distributions

    Multi-Nest [27]
    prop_sigma: 0.01 # Scaling of the spherical gaussian proposal
    N: 2             # Number of proposal distributions
    kde_bw: 0.01     # KDE bandwidth for posterior approximation
\end{verbatim}

\newenvironment{changemargin}[2]{%
\begin{list}{}{%
\setlength{\topsep}{0pt}%
\setlength{\leftmargin}{#1}%
\setlength{\rightmargin}{#2}%
\setlength{\listparindent}{\parindent}%
\setlength{\itemindent}{\parindent}%
\setlength{\parsep}{\parskip}%
}%
\item[]}{\end{list}}

\foreach \method in {gmm,normal,egg}
{
\foreach \x in {1,2,3,4,5,6,7}
{
    \begin{figure*}
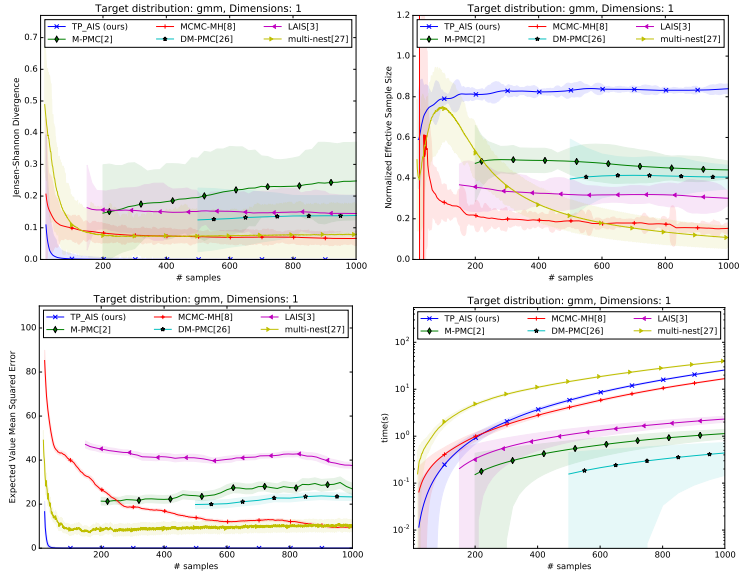

        \centering
        \includegraphics[width=0.4\textwidth]{figures/results/\x_dims_\method_dist_jsd.pdf}
        \includegraphics[width=0.4\textwidth]{figures/results/\x_dims_\method_dist_ness.pdf}
        \includegraphics[width=0.4\textwidth]{figures/results/\x_dims_\method_dist_evmse.pdf}
        \includegraphics[width=0.4\textwidth]{figures/results/\x_dims_\method_dist_time.pdf}
        \caption{Results for \x~dimensions and the \method~target distribution}
    \end{figure*}
}
}

\end{document}